# Control of an Internet-based Robot System Using the Real-time Transport Protocol

Manh Duong PHUNG, Thanh Van Thi NGUYEN, Quang Vinh TRAN

Department of Electronics and Computer Engineering
University of Engineering and Technology
Vietnam National University, Hanoi
Email: duongpm@vnu.edu.vn – vinhtq@vnu.edu.vn

*Abstract*— **In this paper, we introduce a novel approach in controlling robot systems over the Internet. The Real-time Transport Protocol (RTP) is used as the communication protocol instead of traditionally using TCP and UDP. The theoretic analyses, the simulation studies and the experimental implementation have been performed to evaluate the feasibility and effectiveness of the proposed approach for practical uses.**

## I. INTRODUCTION

Since the first Internet-based telerobot appeared in 1994 [1], substantial effort has been devoted to remotely control real-time systems over the Internet, bringing people significant applications such as the virtual laboratory, the telemedicine system and the remote home security system [1]-[5]. It is well recognized that the most challenging and distinct difficulties with these works are associated with the inevitable Internet transmission delays, delay jitter and non-guaranteed bandwidth. To overcome these problems, most of the current works focused on developing advanced remote control algorithms [2]-[4] and interface techniques [5][6]. The data communication between the human operator and the remote robot, however, is usually treated as a given condition and hardly addressed explicitly. For example, most remote control systems directly employ the Transmission Control Protocol (TCP) or the User Datagram Protocol (UDP) as the data transmission protocol even though neither TCP nor UDP was originally designed for real-time applications.

UDP (User Datagram Protocol) is based on the idea of sending a datagram from one device to another as fast as possible without due consideration of the state of the network [7]. This protocol does not maintain a connection between the sender and the receiver, and it does not guarantee that the transmitted data packets will reach the destination as well as the chronological order of the data at the receiving end. In addition, UDP is not equipped with any congestion control mechanism, which means the sending rate cannot be adapted to the real bandwidth available. The main advantage of UDP, however, is the relatively minimized transmission delay and delay jitter achieved under good network conditions.

TCP (Transmission Control Protocol) is a more sophisticated protocol which was originally designed for the reliable transmission of static data such as e-mails and files over low-bandwidth, high-error-rate networks [8]. In each transmission session, TCP establishes a virtual connection between the sender and the receiver, performs the acknowledgment of received data packets, and implements the retransmission mechanism when necessary. TCP can also adapt to the variation of network condition by applying strict congestion control policy with slow start, fast recovery, fast retransmission and window-based flow control mechanisms. With these features, TCP has been very effectively used in the transmission of static data, contributing significantly to the growth of the Internet. However, employing the TCP for real-time applications is hardly appropriate since the timeliness of transmission outweighs the reliability concerns so that the retransmission mechanism is ill-suited for real-time data delivery, and the strict congestion control mechanism incurs higher delay jitter which rapidly degrades the quality of service (QoS) in a congested network.

In this paper, we propose a novel approach in which the Real-time Transport Protocol (RTP) is employed as the transport protocol for controlling an Internet-based robot system. Over the Internet, the operator can control a mobile robot to explore an unknown environment in real time. Many simulations and experiments have been conducted to evaluate the effectiveness and applicability of the proposed approach.

## II. REAL-TIME TRANSPORT PROTOCOL AND REAL-TIME TRANSPORT CONTROL PROTOCOL

Firstly published in 1996, RTP is a relatively new transport protocol but it has since become the standard for delivering real-time multimedia data [9]. There are two parts to RTP: the data transfer protocol called RTP and the associated control protocol called RTCP. While RTP carries the media streams such as audio and video, RTCP is used to monitor transmission statistics and information related to the quality of service.

### A. Real-time Transport Protocol (RTP)

RTP aims to provide services useful for the transport of real-time media, such as audio and video, over IP networks. These services include timing recovery, loss detection and correction, payload and source identification, reception quality feedback, media synchronization, and membership management. RTP provides neither any guarantees for data delivery nor packet delivery in order. The main functions of RTP include:

- Identification of payload type
- Identification of the source sending the RTP packets
- Timestamps to RTP packets
- Sequence numbers to RTP packets

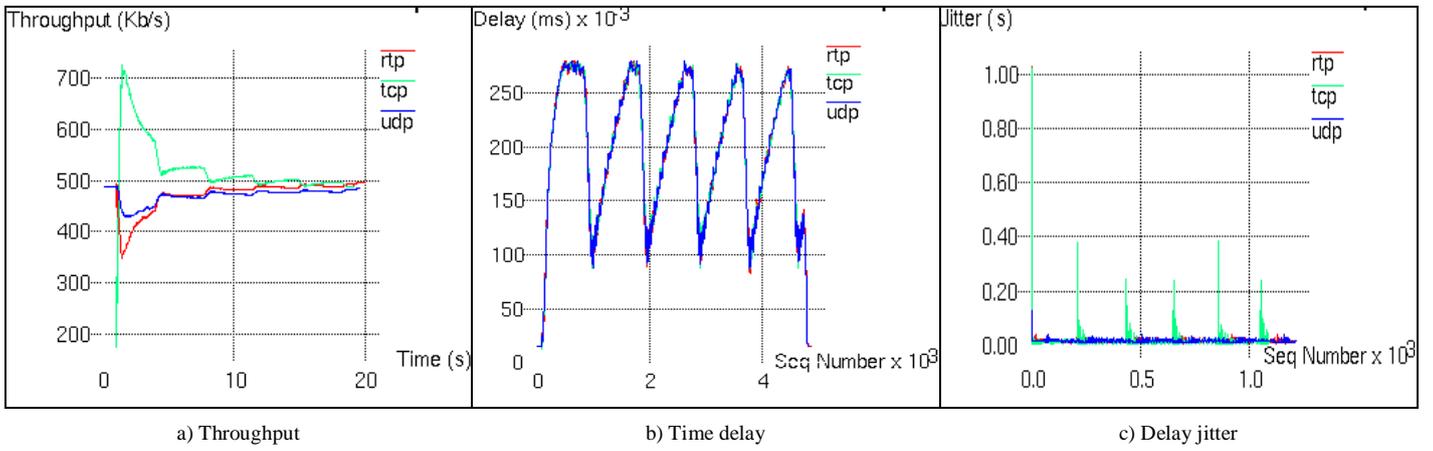

a) Throughput　　　　　　　　　b) Time delay　　　　　　　　　c) Delay jitter

Figure 1: Simulation results in case of no network congestion

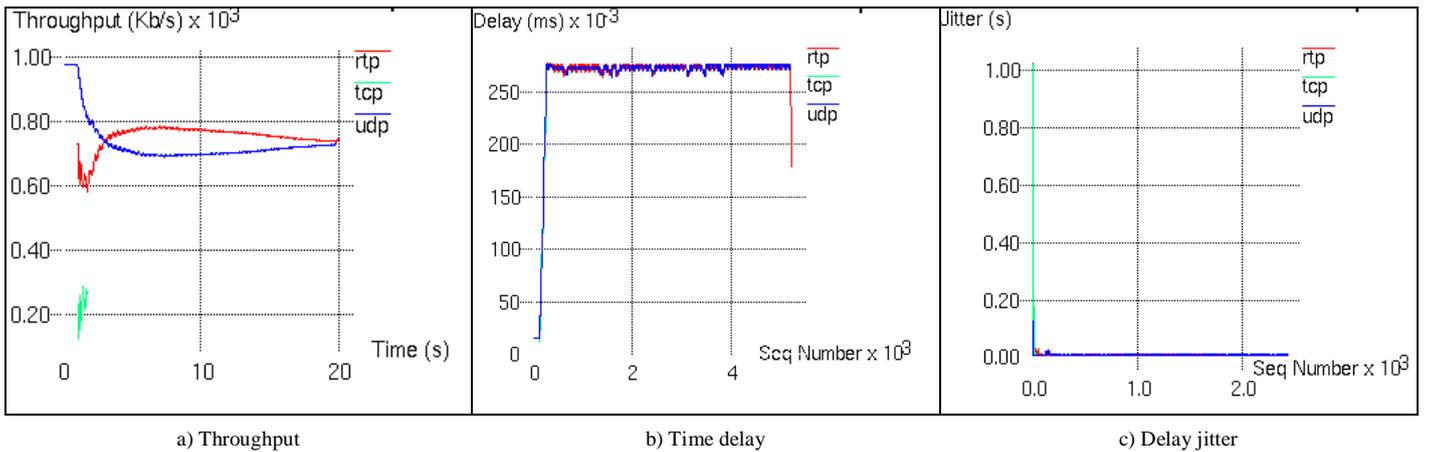

a) Throughput　　　　　　　　　b) Time delay　　　　　　　　　c) Delay jitter

Figure 2: Simulation results in case of network congestion

In the view of real-time system control, some earlier works have suggested that the buffer mechanism of RTP may render it inappropriate for real-time control applications [10][11]. However, in this paper, it will be shown that with an appropriately optimized buffer time and size, the amount of network jitter can be significantly reduced while still satisfying the conditions for system stability.

### B. Real-time Transport Control Protocol (RTCP)

The control protocol, RTCP, is based on the periodic transmission of control packets to all participants in the session, using the same distribution mechanism as the data packets. It performs four functions:

- RTCP provides feedback on the quality of the data distribution such as the packet loss ratio, the delay jitter and the timestamps of sender and receiver reports.
- RTCP carries an identifier of the source sending the packets and other source description information.
- By receiving RTCP packets from all participants, each one can estimate the total number of session participants. This number is used to calculate the rate at which the packets are sent.
- RTCP supplies the information needed to synchronize media streams.

Correct implementation of RTCP can significantly enhance an RTP session: It permits the receiver to lip-sync audio and video, identifies the other members of a session, and allows the sender to make an informed choice of error protection scheme to use to achieve optimum quality.

### III. SIMULATION STUDIES

In this section, we present our simulation studies on RTP to evaluate its network characteristics such as transmission delay, delay jitter and network throughput in comparison with TCP and UDP. We use the widely adopted network simulation tool

*ns*-2, which was developed by Defense Advanced Research Projects Agency (DARPA) through the Virtual InterNetwork Testbed (VINT) project, for the simulations [12]. Two scenarios will be discussed in this section:

- One RTP flow, one TCP flow and one UDP flow share a link without the congestion.
- One RTP flow, one TCP flow and one UDP flow share a bottleneck link under a heavy congestion.

### A. *One RTP flow, one TCP flow and one UDP flow share a link without the congestion*

The network topology of this simulation is shown in fig.3. Three sources of traffic corresponding to the RTP, UDP, and TCP ones are connected to a router. The router forwards the traffic to a sink through a 1.5Mbps duplex link. The RTP and UDP sources send data to the network at 0.5Mbps and the TCP source creates the traffic based on the state of the network.

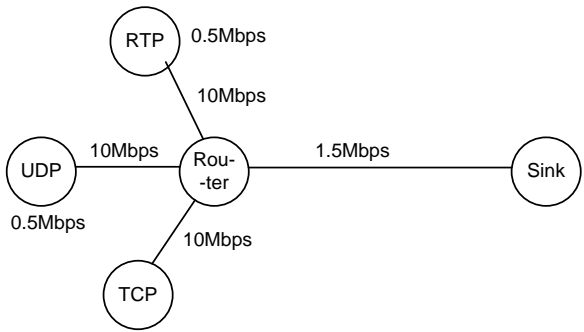

Figure 3: Network topology in the no network congestion simulation

From the results shown in fig.1, we can see that all RTP, UDP and TCP flows share a fair bandwidth of the network and they introduce a common transmission delay behavior. The network jitter of TCP flow is however quite large in comparison with RTP and UDP flows.

### B. *One RTP flow, one TCP flow and one UDP flow share a bottleneck link under a heavy congestion*

The network topology in this simulation (fig.4) is similar to the previous one except that both RTP and UDP flows send the data at 1.0Mbps to the network causing a heavy congestion at the router. Fig.2 shows the simulation results. We can see that the TCP flow cannot compete with RTP and UDP flows to send the traffic to the network and it is dropped from the network.

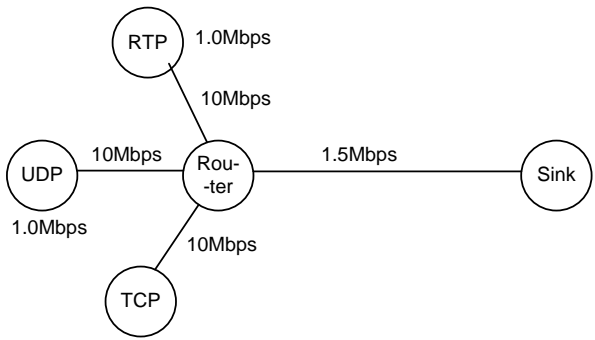

Figure 4: Network topology in the network congestion simulation

In real-time control system, it is important that the data transmission should be maintained at a steady speed while the transmission delay and delay jitter need be reduced to the minimums. From the simulation results, it is well recognized that TCP is not appropriate to real-time control because the delay jitter is relatively high, and the sudden drop in traffic is not acceptable. RTP and UDP show the similar behavior in the simulations. However, as mentioned in the "Introduction" section, UDP is merely a raw protocol that was not intended for practical utilization while RTP was originally designed for transmitting real-time media with the support of RTCP, error concealment and multicast mechanism. The choice of RTP for real-time control therefore is proper.

## IV. EXPERIMENTS

To verify the validity and effectiveness of the proposed approach in control, we implemented an Internet-based robot system in which a mobile robot could be controlled over the Internet using RTP.

### A. *Hardware configuration*

The hardware configuration of the system is shown in fig.5. It consists of three components: the mobile robot, the central server and the client computer. The server has two network interfaces: one to link it to the Internet through a static IP, and the other interface allows it to communicate with the mobile robot via an 802.11a wireless router.

The robot is a commercial Sputnik mobile robot [13]. It has basic components for motor control, sensing and navigation, including battery power, drive motors and wheels, position/speed encoders, infrared sensors, integrated sonar ranging sensors and a visual system. Sensing and motor control are managed by an on-board digital signal processor (DSP) with an independent motor/power and sonar controller boards for a versatile operating environment. The drive system uses high-speed, high-torque, reversible-DC motors. The Sputnik mobile robot provides three sonar sensors. One is mounted to the front of the robot and the others are mounted to its left and right side respectively. The visual system is detachable and mounted on the head of the Sputnik mobile robot. It mainly consists of a MCI3908 color image module with Sharp mini color camera head LZ0P390M. The image size can be up to CIF format (353 x 288 pixels), and the operation frame rate is up to 15 fps.

### B. *System Software*

The system software employs client-server architecture for robot control and feedback information display (fig.6). In our system, the server handles all control requests from the client, processes them, and forwards translated commands to the Sputnik robot. The server also retrieves sensor data of the robot and transmits it to clients. In short, the server acts as the bridge between the client and the Sputnik mobile robot. The server program is written in JAVA [14] and the communication protocol between the server and client is RTP.

On the client side, the graphical user interface (GUI) is responsible for receiving and interpreting human control commands such as mouse-click events in the control panel, and transmitting them to the server program by RTP. The GUI also displays feedback information about the remote side such as current robot position, ultrasonic ranging values, and live video of the remote environment.

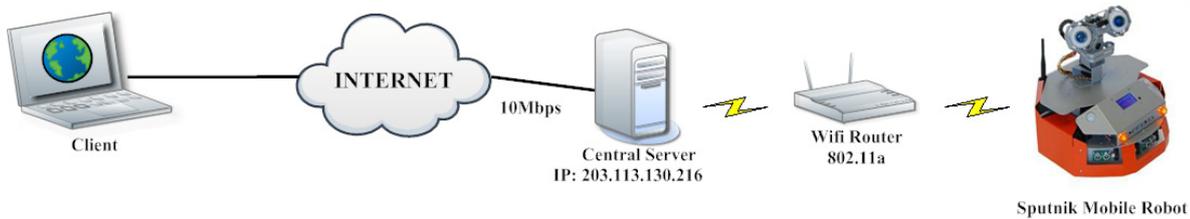

Figure 5: System hardware configuration

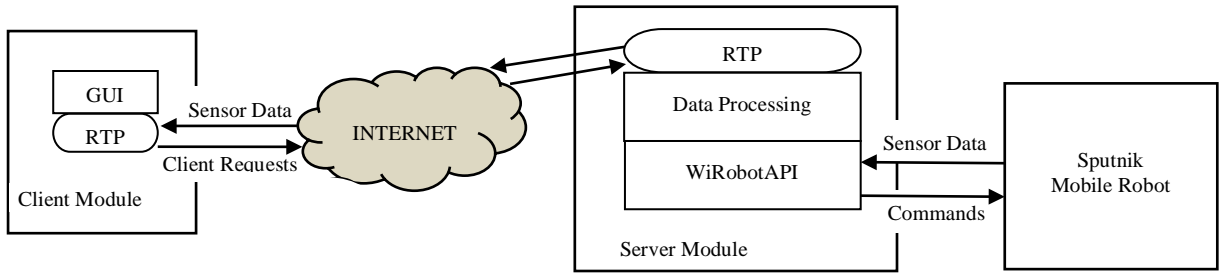

Figure 6: System software architecture

## C. Experimental results

To evaluate the performance of the system, we have carried out many experiments. Fig.7 shows the setup of the environment that the Sputnik mobile robot moves through. The robot is located at the Automatic Control and Robotics Laboratory (ACRs Lab) of the Hanoi University of Engineering and Technology, Vietnam National University, and it is controlled over the Internet by an operator who is 12km far away. The average speed of the robot is 10cm/s. The goal of the experiments is to remotely guide the Sputnik robot from the starting point $O_o$ to the objective point $O_d$.

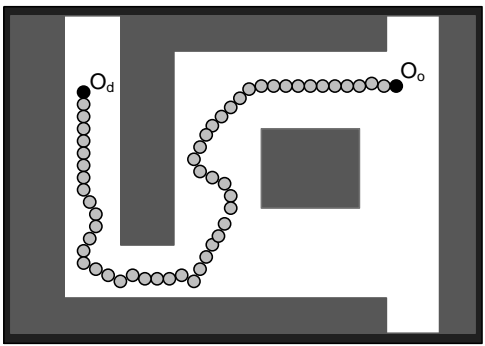

Figure 7: Remotely navigating the mobile robot around the laboratory

In the experiments, by using the designed client GUI, the user successfully navigated the Sputnik mobile robot from the point $O_o$ to the point $O_d$ via the Internet (Fig.7). This result can be verified from the fig.8 and fig.9: an average delay of 43ms and an average jitter of 4ms definitely satisfy the stable conditions of this particular setup which can tolerate the delay and jitter up to several seconds. It is also noted from fig.9 that an average jitter of 4ms is much better than the 15ms jitter requirement for normal video streaming [15].

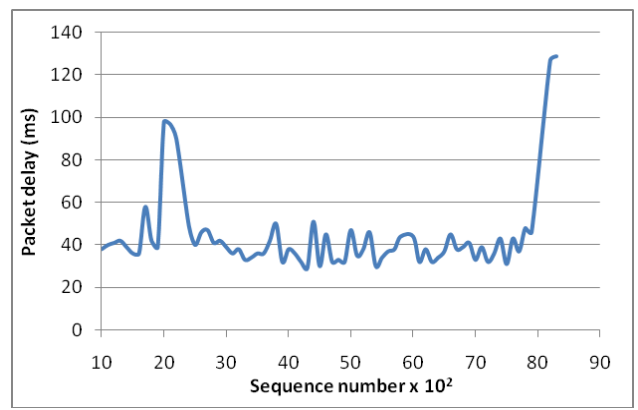

Figure 8: Packet delay during the experiment

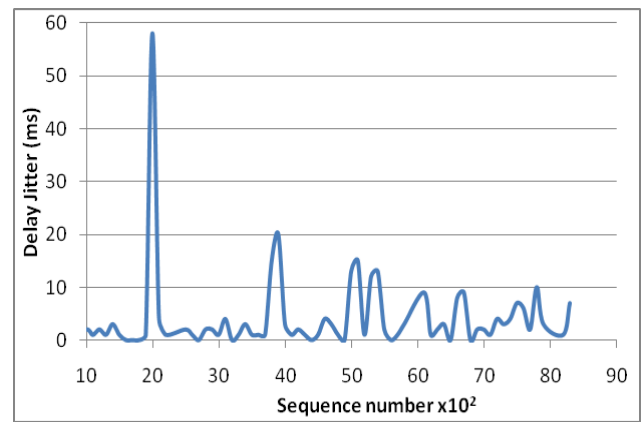

Figure 9: Delay jitter during the experiment

Fig.10 shows a sequence of the snapshots of the Sputnik mobile robot when it was remotely being guided via the Internet to move from point $O_o$ to point $O_d$ in the ACRs Lab.

We have conducted the experiments in different day times- morning, noon, afternoon and night-trying to capture the ''rush hour'' of the Internet traffic. At all times, the user succeeded to navigate the mobile robot through the ACRs Lab.

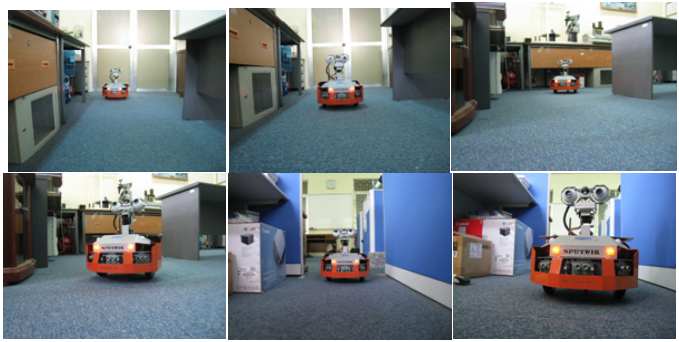

Figure 10: A sequence of images showing the motion of robot in a laboratory environment during the tele-operation

## V. CONCLUSIONS

In this paper, we propose a novel approach in controlling Internet-based robot systems: the use of the Real-time Transport Protocol. The core part of this approach, RTP, is analyzed theoretically and its performances, such as time delay and network throughput, are evaluated by using simulation studies. The results confirm the theoretical analysis very well. Furthermore, the experimental implementation on an Internet mobile robot shows that the introduced scheme provides acceptable delay and jitter for the real-time control.